\documentclass[10pt,twocolumn,letterpaper]{article}

\usepackage{cvpr}              

\usepackage{graphicx}
\usepackage{amsmath}
\usepackage{amssymb}
\usepackage{booktabs}
\usepackage{svg}

\usepackage[pagebackref,breaklinks,colorlinks]{hyperref}
\usepackage[marginal]{footmisc}

\usepackage[capitalize]{cleveref}
\crefname{section}{Sec.}{Secs.}
\Crefname{section}{Section}{Sections}
\Crefname{table}{Table}{Tables}
\crefname{table}{Tab.}{Tabs.}
\crefname{table}{Tab.}{Tabs.}

\usepackage{xcolor}
\usepackage{enumitem}
\usepackage{xspace}
\usepackage{booktabs}
\usepackage{colortbl}
\usepackage{multirow}
\usepackage{makecell}
\usepackage{lipsum} 
\usepackage{gensymb} 

\newcommand{\Tref}[1]{Table~\ref{#1}}
\newcommand{\eref}[1]{Eq.~\eqref{#1}}

\newcommand{\fref}[1]{Fig.~\ref{#1}}
\newcommand{\Fref}[1]{Figure~\ref{#1}}
\newcommand{\sref}[1]{Sec.~\ref{#1}}

\newcommand{\methodnameabbr}{REC-MV\xspace}

\newcommand{\canop}{\mathbf{p}}

\newcommand{\image}{I}

\newcommand{\camera}{$\pi$\xspace}
\newcommand{\sdfws}[1]{$\mathbf{\eta}_{#1}$\xspace}
\newcommand{\msdfws}[1]{\mathbf{\eta}_{#1}\xspace}

\newcommand{\curvetwo}{\zeta}

\newcommand{\deformfield}{\Phi}
\newcommand{\deformMLP}{\mathcal{D}}
\newcommand{\smplweights}{\boldsymbol{\theta}}
\newcommand{\smplshape}{\boldsymbol{\beta}}
\newcommand{\skinning}{\mathcal{W}}

\newcommand{\defws}{\phi}
\newcommand{\deflant}{\mathbf{h}}
\newcommand{\pos}[1]{E(#1)}

\newcommand{\rendermlp}{f_c}
\newcommand{\canon}{\mathbf{n}}
\newcommand{\renderws}{\psi}
\newcommand{\renderlant}{\mathbf{z}}
\newcommand{\cameraview}{\textbf{v}}

\newcommand{\pcolors}[1]{C_{#1}}

\newcommand{\tmpcurve}{\mathbf{L}}
\newcommand{\curve}{{\mathcal{C}}}

\newcommand{\inic}[1]{i}
\newcommand{\predc}[1]{\curve(#1)}
\newcommand{\centerc}{\mathbf{p_c}}
\newcommand{\tradline}{\mathbf{L}}
\newcommand{\scaleline}{\mathbf{\bar{L}}}

\newcommand{\projf}{\mathrm{CD}}

\newcommand{\ndir}[1]{\mathbf{n^d_{#1}}}
\newcommand{\nsdir}[1]{\mathbf{n^c_{#1}}}
\newcommand{\sndir}[1]{S^d_{#1}}
\newcommand{\snsdir}[1]{S^c_{#1}}
\newcommand{\alphags}{s}
\newcommand{\alphagt}{\mathbf{t}}
\newcommand{\alphagR}{\mathbf{R}}

\newcommand{\surf}[1]{S_{#1}}
\newcommand{\canos}{\mathbf{T_s}}
\newcommand{\updatesur}{\mathbf{\hat{T}_s}}
\newcommand{\curvesurf}{\mathbf{T_\curve}}

\newcommand{\loss}{\mathcal{L}}
\newcommand{\lossproj}{\loss_{proj}}
\newcommand{\lossslop}{\loss_{slop}}
\newcommand{\lossanap}{\loss_{anap}}
\newcommand{\losscurve}{\loss_{curve}}
\newcommand{\lossrgb}{\loss_{RGB}}
\newcommand{\lossnorm}{\loss_{norm}}
\newcommand{\lossrigid}{\loss_{arap}}
\newcommand{\losseik}{\loss_{eik}}
\newcommand{\lossconss}{\loss_{mcons}}
\newcommand{\lossconsc}{\loss_{ccons}}
\newcommand{\lossims}{\loss_{ims}}

\newcommand{\lprojw}{\lambda_{proj}}
\newcommand{\lslopw}{\lambda_{slop}}
\newcommand{\lanap}{\lambda_{anap}}
\newcommand{\surfW}{\mathbf{\Theta}}

\newcommand{\larapw}{\lambda_{arap}}
\newcommand{\lnormw}{\lambda_{norm}}
\newcommand{\leikw}{\lambda_{eik}}

\newcommand{\lconss}{\lambda_{mcons}}
\newcommand{\lconsc}{\lambda_{ccons}}

\newcommand{\Frst}[1]{\textbf{#1}}

\usepackage[labelsep=period]{caption}
\renewcommand{\paragraph}[1]{\vspace{0.2em}\noindent \textbf{#1}}

\definecolor{MyDarkRed}{rgb}{0.66, 0.16, 0.16}
\definecolor{MyDarkBlue}{rgb}{0.16, 0.16, 0.66}
\usepackage{lipsum} 

\newcommand{\wlink}[1]{\textcolor{magenta}{{#1}}}

\def\cvprPaperID{2001} 
\def\confName{CVPR}
\def\confYear{2023}

\begin{document}

  
\title{REC-MV: REconstructing 3D Dynamic Cloth from Monocular Videos}

\author{Lingteng Qiu$^{1}\footnotemark[1]$ \quad Guanying Chen$^{1,2}\footnotemark[1]$ \quad Jiapeng Zhou$^{1}$ \quad Mutian Xu$^{1}$ \\ \quad Junle Wang$^{3}$ \quad Xiaoguang Han$^{1,2}\footnotemark[2]$ \vspace{0.3em} \\
{\normalsize $^1$SSE, CUHKSZ}
\quad{\normalsize $^2$FNii, CUHKSZ} \quad {\normalsize $^3$Tencent}
}


\maketitle

\footnotetext[1]{Equal contribution.}
\footnotetext[2]{Corresponding author: \wlink{hanxiaoguang@cuhk.edu.cn}.}


\begin{abstract}
    Reconstructing dynamic 3D garment surfaces with open boundaries from monocular videos is an important problem as it provides a practical and low-cost solution for clothes digitization.
    Recent neural rendering methods achieve high-quality dynamic clothed human reconstruction results from monocular video, but these methods cannot separate the garment surface from the body.
    Moreover, despite existing garment reconstruction methods based on feature curve representation demonstrating impressive results for garment reconstruction from a single image, they struggle to generate temporally consistent surfaces for the video input.
    To address the above limitations, in this paper, we formulate this task as an optimization problem of 3D garment feature curves and surface reconstruction from monocular video.
    We introduce a novel approach, called \textbf{\emph{\methodnameabbr}}, to jointly optimize the explicit feature curves and the implicit signed distance field (SDF) of the garments. Then the open garment meshes can be extracted via garment template registration in the canonical space.
    Experiments on multiple casually captured datasets show that our approach outperforms existing methods and can produce high-quality dynamic garment surfaces. The source code is
    available at \wlink{https://github.com/GAP-LAB-CUHK-SZ/REC-MV}.


\end{abstract}

\section{Introduction}
\label{sec:intro}
High-fidelity clothes digitization plays an essential role in various human-related vision applications such as virtual shopping, film, and gaming. 
In our daily life, humans are always in a moving status, driving their clothes to move together. To realize this very common scenario, it is indispensable to gain dynamic garments in real applications.
Thanks to the rapid development of mobile devices in terms of digital cameras, processors, and storage, shooting a monocular video in the wild becomes highly convenient and accessible for general customers.
In this paper, our goal is definite -- \textit{extracting \textbf{dynamic 3D garments} from \textbf{monocular videos}}, which is significantly meaningful and valuable for practical applications, but is yet an uncultivated land with many challenges.

\begin{figure}[tb] \centering
   \includegraphics[width=0.48\textwidth,height=0.3\textwidth]{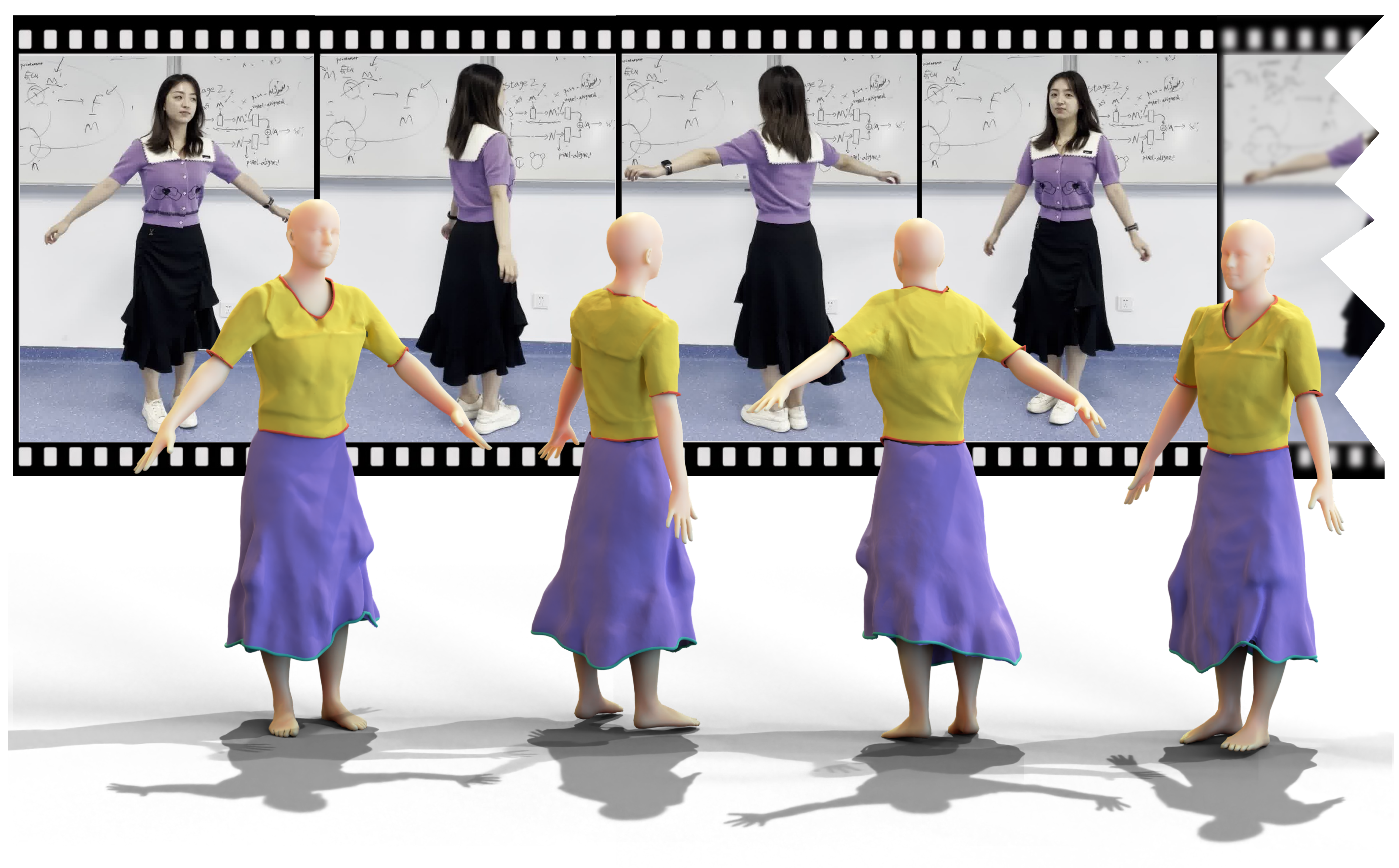}\\
   \vspace{-0.5em}
   \caption{\textit{Can we extract \textbf{dynamic 3D garments} from \textbf{monocular videos}}? The answer is Yes! By jointly optimizing the dynamic feature curves and garment surface followed by non-rigid template registration, our method can reconstruct high-fidelity and temporally consistent garment meshes with open boundaries.} \label{fig:teaser}
\end{figure}

We attempt to seek a new solution to this open problem and start by revisiting existing works from two mainstreams.
i) 
Leveraging the success of neural rendering  methods~\cite{mildenhall2020_nerf_eccv20,yariv2020multiview,niemeyer2020differentiable}, several works are able to reconstruct dynamic clothed humans from monocular videos \cite{su2021nerf,chen2021animatable,weng2022humannerf,li2022avatarcap,jiang2022selfrecon}, by representing the body surface with an implicit function in the canonical space and apply skinning based deformation for motion modeling.
One naive way to achieve our goal is: first to get the clothed human through these methods and separate the garments from human bodies. However, such a separation job requires laborious and non-trivial processing by professional artists, which is neither straightforward nor feasible for general application scenarios.
ii) 
As for garment reconstruction, many methods \cite{bhatnagar2019multi,jiang2020bcnet,zhu2020deep,corona2021smplicit,zhu2022registering} make it possible to reconstruct high-quality garment meshes from single-view images in the wild. 
Specifically, ReEF~\cite{zhu2022registering} estimates 3D feature curves\footnote{feature curves of the garment (\eg, necklines, hemlines) can provide critical cues for determining the shape contours of the garment.} and an implicit surface field~\cite{mescheder2019occupancy} for non-rigid garment template registration.
Nonetheless, these methods struggle to produce temporally consistent surfaces when taking videos as inputs. 
The above discussion motivates us to combine the merits of both the dynamic surface modeling in recent neural rendering methods and the explicit curve representation for garment modeling. 
To this end, we try to delineate a new path towards our goal: \textit{optimizing dynamic explicit feature curves and implicit garment surface from monocular videos}, to extract temporally consistent garment meshes with open boundaries.
We represent the explicit curves and implicit surface in the canonical space with skinning-based motion modeling, and optimize them by 2D supervision automatically extracted from the video (\eg, image intensities, garment masks, and visible feature curves). After that, the open garment meshes can be extracted by a garment template registration in the canonical space (see~\fref{fig:teaser}).

We strive to probe this path as follows:
\textbf{(1)} As a feature curve is a point set whose deformation has a high degree of freedom, directly optimizing the per-point offsets often leads to undesired self-intersection and spike artifacts. To better regularize the deformation of curves, we introduce an \emph{intersection-free curve deformation} method to maintain the order of feature curves. 
\textbf{(2)} We optimize the 3D feature curves using 2D projection loss measured by the estimated 2D visible curves, where the key challenge is to accurately compute the visibility of curves. To address this problem, we propose a \emph{surface-aware curve visibility estimation} method based on the implicit garment surface and z-buffer.
\textbf{(3)} To ensure the accuracy of curve visibility estimation during the optimization process, the curves should always be right on the garment surface. We therefore introduce a \emph{progressive curve and surface evolution} strategy to jointly update the curves and surface while imposing the on-surface regularization for curves.

To summarize, the main contributions of this work are:
\begin{itemize}[itemsep=0pt,parsep=0pt,topsep=2bp]
    \item We introduce \textbf{\methodnameabbr}, to our best knowledge, the \textbf{first} method to reconstruct dynamic and open loose garments from the monocular video.
    \item We propose a new approach for joint optimization of explicit feature curves and implicit garment surface from monocular video, based on carefully designed intersection-free curve deformation, surface-aware curve visibility estimation, and progressive curve and surface evolution methods. 
    \item Extensive evaluations on casually captured monocular videos demonstrate that our method outperforms existing methods. 
\end{itemize}

\section{Related Work}
\label{sec:related_works}


\paragraph{Human Reconstruction from Single-view Image.} 
Traditional methods for human reconstruction often adopt a parametric human model (\eg, SMPL~\cite{loper2015smpl} or SCAPE~\cite{anguelov2005scape}) and can only recover a naked 3D body~\cite{hmrKanazawa17,joo2018total}.
To increase the surface details, free-form deformations can be applied to the mesh vertices to model small geometry variations caused by the clothing~\cite{alldieck2018video,alldieck19cvpr,lazova3dv2019,generativeclothing,monoclothcap}.

Recent methods propose to utilize implicit surface representations~\cite{park2019deepsdf, mescheder2019occupancy} to reconstruct 3D clothed human with an arbitrary topology.
Specifically, PIFu and PIFuhd~\cite{saito2019pifu,saito2020pifuhd} extract pixel-aligned spatial features from images as the input for implicit surface function for occupancy prediction. Follow-up methods then integrates 3D-aligned features to improve the results~\cite{huang2020arch,he2021arch++,zheng2020pamir,he2020geo,cao2022jiff,yang2021s3,hong2021stereopifu}.
As these methods only consider single-image reconstruction, they cannot produce temporally consistent results for video input.


\paragraph{Human Reconstruction from Monocular Video.} 
Inspired by the success of neural rendering methods~\cite{mildenhall2020_nerf_eccv20,yariv2020multiview,niemeyer2020differentiable} in scene reconstruction, many methods have been proposed to reconstruct 3D human from sparse-view~\cite{peng2021neural,zhao2021humannerf,liu2021neural,xu2021h,yang2022banmo} or monocular~\cite{su2021nerf,weng2022humannerf,jiang2022selfrecon} videos.

Anim-NeRF~\cite{chen2021animatable}, Neuman~\cite{jiang2022neuman} and HumanNeRF~\cite{weng2022humannerf} introduce methods based on neural radiance field (NeRF)~\cite{mildenhall2020_nerf_eccv20} to reconstruct an animatable avatar from monocular video. These methods transform a 3D point in the observation space to the canonical space by inverse-skinning, and then perform volume rendering in the canonical space.
A-NeRF~\cite{su2021nerf} additionally adopt a skeleton-relative encoding strategy.
AvatarCap \cite{li2022avatarcap} proposes a monocular human volumetric capture method, but requires reconstructing an avatar from multiple 3D scans in advance.



\begin{figure*}[t] \centering
    \includegraphics[width=\textwidth]{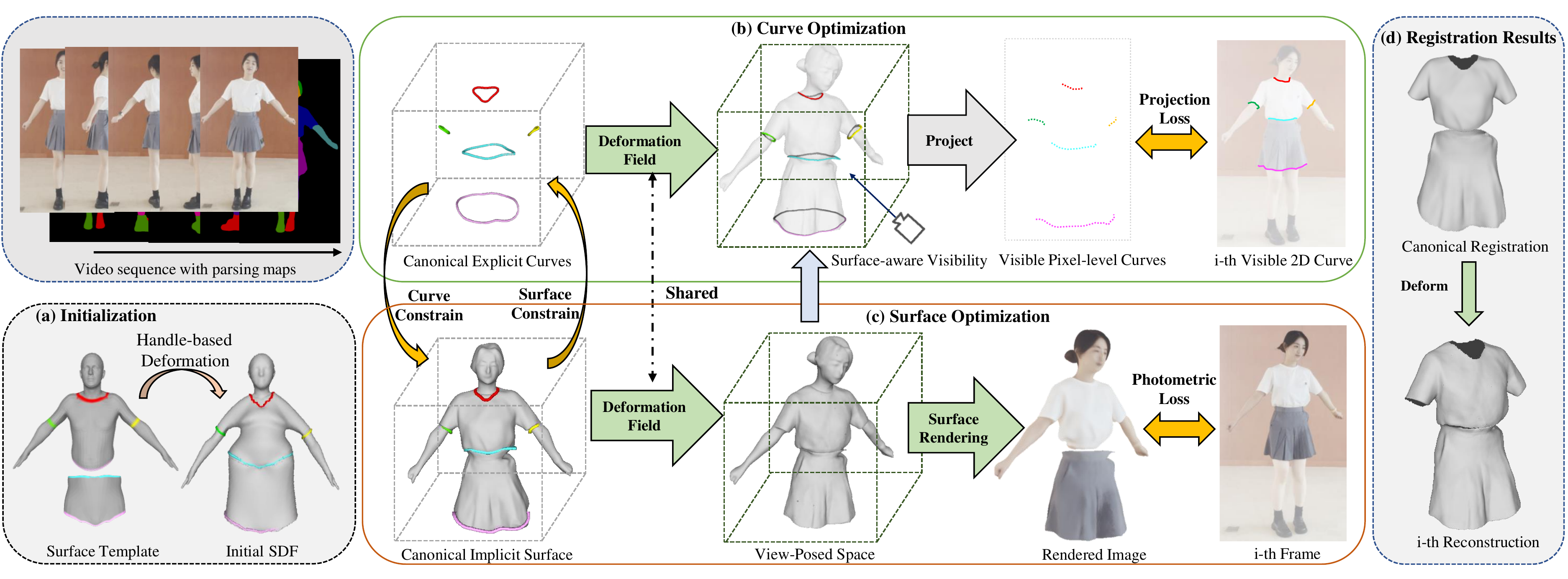} \\
    \caption{
        \textbf{Overview of the proposed \methodnameabbr}. 
    (a) Starting from a surface template, we initialize the canonical curves by solving \eref{eq:sRT}, and apply a handle-based deformation to initialize the canonical implicit surface.
    (b) Given an $i$-th frame, canonical curves are deformed to the camera view space to compute the projection loss based on the surface-aware visibility estimation.
    (c) Similarly, the canonical surface is deform to the camera view to compute the photometric loss by differentiable rendering.
    The curves and surface are jointly optimized to enable a progressive co-evolution.
    (d) Last, the open garment meshes can be extracted by template registration in the canonical space.
    } \label{fig:pipeline}
\end{figure*}

\paragraph{Garment Reconstruction from Images.}
Reconstructing garment mesh from images enables many applications like virtual try-on and content creation. Existing methods reconstruct the clothing as a separate layer on top of the body~\cite{ponsmollSIGGRAPH17clothcap,lahner2018deepwrinkles,jin2018pixel,tiwari20sizer,xiang2022dressing, casado2022pergamo}.
Among them, several methods address the challenging problem of garment reconstruction from single-view image~\cite{bhatnagar2019multi,jiang2020bcnet,corona2021smplicit,moon20223d,zhu2020deep,zhu2022registering}.
MGN~\cite{bhatnagar2019multi} learns a per-category parametric model from a large-scale clothing dataset. 
BCNet~\cite{jiang2020bcnet} first reconstructs a coarse template and then refines the surface details with a displacement network.  
AnchorUDF~\cite{zhao2021learning} adopts the unsigned distance field~(UDF)~\cite{chibane2020ndf} to represent the open surface mesh.
SMPlicit~\cite{corona2021smplicit} proposes a generative model to reconstruct layered garments from a single image. 
DeepFasion3D~\cite{zhu2020deep} reconstructs the surface with occupancy network~\cite{mescheder2019occupancy} and applys non-rigid ICP to register the clothing template.
ReEF~\cite{zhu2022registering} registers explicit clothing template to the implicit field learned from pixel-aligned implicit function. However, as these single-image methods do not consider clothing motion, they are not suitable for dynamic garment reconstruction.

Among methods related to garment reconstruction from videos,
Li~\etal~\cite{li2021deep} introduce a method to learn physics-aware clothing deformation from monocular videos, but assumes the template scans for the body and clothing are provided~\cite{habermann2020deepcap}.
Garment Avatar~\cite{halimi2022garment} proposes a multi-view patterned cloth tracking algorithm, requiring the subject to wear clothing with specific patterns.
SCARF represents the layered clothing using radiance field~\cite{feng2022capturing} on top of the SMPL-X model~\cite{pavlakos2019expressive} from monocular video. 
In contrast, our method first reconstructs the explicit 3D garment curves and surfaces, and then extracts the garment mesh via template registration.

\section{Method}%
\label{sec:method}

Given a monocular video with $N_i$ frames depicting a moving person $\{\image_t| t = 1, \dots, N_i\}$, \methodnameabbr aims to reconstruct high-fidelity and space-time coherent \emph{open} garment meshes. 
This is a challenging problem as it requires a method to simultaneously capture the shape contours, local surface details, and the motion of the garment. 

Observing that feature curves (\eg, necklines, hemlines) provide critical cues for determining the shape contours of garment~\cite{zhu2020deep} and implicit signed distance function (SDF) can well represent a detailed \emph{closed} surface~\cite{jiang2022selfrecon}, 
we propose to first optimize the explicit 3D feature curves and implicit garment surfaces from the video, and then apply non-rigid clothing template registration to extract the open garment meshes (see~\fref{fig:pipeline}).

\paragraph{Preprocessing.}
We generate the initial shape parameter $\smplshape$, camera intrinsic \camera, and per-frame SMPL~\cite{loper2015smpl} pose parameters \{$\smplweights_t |t = 1, \dots , N_i$\} using Videoavatar~\cite{alldieck2018video}.
To identify the garment regions in 2D images, we apply the existing garment parsing method~\cite{li2020self} to estimate the garment masks.
Our method also requires 2D visible curves~$\zeta=\{\zeta_{l,t}| l=1,\dots N_l, t = 1,\dots, N_i\}$ for 3D curve recovery, where $N_l$ denotes the number of curves. Note that the 2D visible curves can be automatically produced by parsing boundaries of the garment mask (more details in the supplementary material). 

\paragraph{Overview.} To utilize the information that exists in the entire video for dynamic garment reconstruction, we represent the explicit feature curve and implicit garment surface in the canonical space~(\sref{sub:curve and sdf}).
For a specific time step, we adopt the skeleton-based skinning and non-rigid deformation modeling to map the canonical curves and surfaces to the camera view space~(\sref{sub:skinning}).
As the given 2D curves only contain visible points, to optimize the 3D feature curves from 2D projection error, we propose a surface-aware approach to compute the visibility of the 3D feature curve based on z-buffer (\sref{sub: curve capture}).
In terms of implicit surface optimization, we minimize the photometric loss between the rendered and input image based on the differentiable surface rendering technique (\sref{sub: surface capture}). 
Then the adopted loss functions for joint optimization of curves and surfaces are described (\sref{sub: loss}).
Last, the open garment meshes can be extracted by registering an explicit garment template to the recovered curves and implicit surfaces in the canonical space (more details in the supplementary material).
Then the garment meshes can be deformed based on the SMPL poses.

\subsection{Feature Curve and Surface Representation}%
\label{sub:curve and sdf} 

\paragraph{Explicit Surface Template.} 
Following DeepFashion3D~\cite{zhu2020deep}, we employ several surface templates, each contains a pre-defined set of 3D feature curves $\tmpcurve=\{\tmpcurve_i|i=1, \dots, N_{l}\}$ extracted from the garment boundaries, where $N_l$ is the number of feature curves (see our supplementary materials for more details). 
The surface templates will be used for garment surface initialization and the pre-defined feature curves will be used for curve initialization\footnote{including templates for uppers, dresses, coats, pants, and skirts.}.

\paragraph{Intersection-free Curve Deformation.} 
A straightforward idea is to represent a feature curve as a discrete point set, and directly estimate the 3D deformation offset for each point during optimization.
However, this unstructured curve representation struggles to maintain the order of the points and often generate spike artifacts due to the high degree of freedom of the deformation.

\begin{figure}[tb] \centering
    \includegraphics[width=0.4\textwidth]{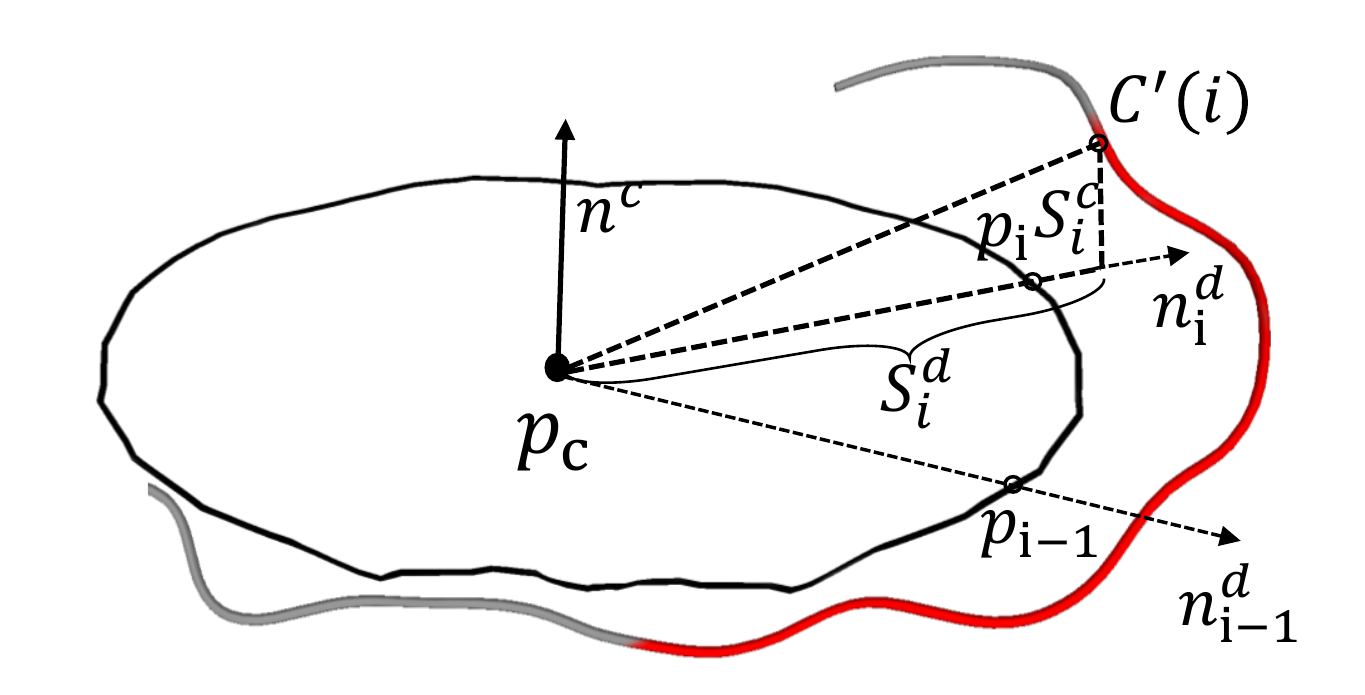}
    \caption{Illustration of the intersection-free curve deformation.} \label{fig:curve}
\end{figure}

To address this issue, we introduce a novel intersection-free curve deformation method, in which the point's deformation at each step is controlled by the curve center and two orthogonal directions ~(see~\fref{fig:curve} for illustration). 
Formally, given a curve $\curve$ of $N_p$ points with center $\centerc$, the updated position of $i$-th point $\curve(i)$ is defined as
\begin{equation}
\begin{aligned}
&\curve'(i) = \centerc + \sndir{i}\ndir{i} +\snsdir{i}\nsdir{}, 
\end{aligned}
\end{equation}
where $\ndir{i}$ is the direction from the curve center to the current point $\curve(i)$, and $\nsdir{}= \frac{1}{N_p-1}\sum_{i=1}^{N_p}(\ndir{i}\times\ndir{i-1})$ is the direction perpendicular to the current feature curve plane.
$\sndir{i}\in \mathbb{R}$ and $\snsdir{i}\in \mathbb{R}$ are learnable parameters specifying the step size of the deformation. 

The proposed intersection-free curve deformation can well preserve the order of points in the curve, which largely reduced the difficulty of optimization compared to the direct offset estimation approach.


\paragraph{Implicit SDF in Canonical Space.} 
Unsigned distance field~(UDF)~\cite{chibane2020ndf} is an implicit function that can represent an open surface. 
However, as UDF is not differentiable at points close to the surface, it is non-trivial to integrate UDF with differentiable surface rendering to take advantage of supervision from 2D photometric loss.
We therefore adopt the SDF to represent a closed garment surface for surface geometry recovery, followed by garment template registration to extract the open surface.

It is common to represent the whole surface with a single SDF for human reconstruction~\cite{jiang2022selfrecon}. 
However, as our goal is to reconstruct separate clothes, using a single SDF to represent both the upper clothes and bottom clothes (\eg, skirt) increases the difficulty of template registration (\ie, splitting the upper and bottom clothes requires highly accurate waist curves). 

To enable better template registration, we consider three different surface types (\ie, \emph{upper-clothing}, \emph{bottom-clothing}, and \emph{upper-bottom}) according to the garment types, and represent each surface type as the zero-isosurface of an independent SDF in the canonical space. The SDF is expressed by an MLP ${f}$ with learnable weights \sdfws:
\begin{equation*}
    \begin{aligned}
     \surf{}(\msdfws{}) = \{\canop \in \mathbb{R}^{3}|f(\canop;\msdfws{}) =0\}.
    \end{aligned}
\end{equation*}

For the sake of simplicity and without loss of generality, 
we illustrate our method in reconstructing a single surface type later in this section.


\subsection{Skinning Based Motion Modeling}
\label{sub:skinning}
We model large body motions by linear blend skinning (LBS) transformation based on the SMPL~\cite{loper2015smpl} model, and utilize a non-rigid deformation field to account for fine-grained deformations.

\paragraph{Skinning Transformation.} 
Given a SMPL body with shape parameter $\smplshape$ and a pose parameter $\smplweights_i$ in $i$-th frame, a point $\canop$ on the body surface in canonical space with skinning weights $w(\canop)$ can be warped to camera view space via skinning transformation $\skinning$.

Notably, the skinning weights $w(\canop)$ are only defined for points on the SMPL surface. To warp arbitrary points in the canonical space to camera view, we use the diffused skinning strategy~\cite{lin2022learning} to propagate the skinning weights of SMPL body vertices to the entire canonical space, and store the weights in a voxel grid of size $256 \times 256 \times 256$. Then we can obtain the skinning weights by trilinear interpolation.

\paragraph{Non-rigid Deformation.} Skinning deformation enables the garment surface to deform in a way consistent with the body's large-scale motion~\cite{he2021arch++}. However, the motion of details and garment parts that are far away from body cannot be fully represented by skinning transformation~\cite{jiang2022selfrecon}. 
Hence, a non-rigid deformation MLP is used to model these fine-grained changes. 
Specifically, we design an MLP $\deformMLP$ with learnable parameters $\defws$ to model garment surface's non-rigid deformation:
\begin{equation}
    \begin{aligned}
     \canop' &= \deformMLP(\canop, \deflant, \pos{\canop};\defws),
    \end{aligned}
\end{equation}
where $\canop'$ is the deformed point of the input point $\canop$ in the canonical space, $\deflant$ is the latent code of the current frame, and $\pos{\canop}$ of $\canop$ is the position encoding~\cite{mildenhall2020_nerf_eccv20} to represent the high-frequency information of spatial points.

Finally, combining $\deformMLP$ with skinning transformation field $\skinning$, we could define a deformation field $\deformfield(\cdot) = \skinning(\deformMLP(\cdot))$ to warp any points in the canonical space to the camera view.

\subsection{3D Feature Curves from 2D Projections}
\label{sub: curve capture}
The 3D feature curve will be optimized by minimizing the distance between its 2D projection on the image plane and the provided 2D visible curves. The key challenge here is how to compute the visibility of the 3D curves in the camera view. 
We first introduce a curve initialization strategy based on rigid transformation, and then propose a surface-aware curve visibility estimation method to support accurate non-rigid curves optimization.

\paragraph{Feature Curve Initialization.} 
We start from the predefined feature curve sets $\tmpcurve=\{\tmpcurve_i|i=1, \dots, N_{l}\}$ provided in the garment template.
To reduce the difficulty of curve optimization, we perform a rigid curve initialization by directly minimizing the Chamfer Distance (CD) between the projected curves on the camera view space and the corresponding visible 2D curves $\curvetwo$ as 
\begin{align}
    \alphags,\alphagt,\alphagR &= \mathop{\arg\min}_{\alphags,\alphagt,\alphagR} \projf\left(\Pi\left(\skinning(\scaleline_i)\right), \curvetwo_i\right), \label{eq:sRT} \\
    \scaleline_i & = \alphags\alphagR(\tradline_i) +\alphagt, \label{eq:initial_curve} 
\end{align}
where $\Pi$ is the projection matrix, $\scaleline_i$ is the transformed feature curve. 
$\alphagt \in \mathbb{R}^3$, $\alphagR\in SO(3)$, and $\alphags\in \mathbb{R} $ are the optimized translation, rotation, and scaling parameters, respectively.

In our implementation, we execute $150$ gradient descent iterations to solve the rigid transformation parameters. 
After rigid optimization, we set $\scaleline$ as the initial position for the feature curve sets $\{\curve_i|i=1,\dots,N_l\}$ for later non-rigid optimization. 

\paragraph{Surface-aware Curve Visibility Estimation.}
As the 2D feature curve $\curvetwo$ only contains visible points, it is essential to identify the visible points of the 3D curve $\curve$ in camera space. 
A naive solution is to consider a point $\predc{i}$ as visible if the cosine similarity between the view direction $\cameraview$ and $\ndir{i}$ (\ie, the direction from curve center to the $i$-th point) in view-pose is less than $0$. 
However, this approach will produce wrong judgments when a curve is occluded by other body parts. 

To tackle this problem, a surface-aware curve visibility estimation method is proposed. 
Specifically, we generate an explicit mesh $\canos$ from implicit surface $\surf{}(\msdfws{})$ in canonical space via marching cube~\cite{lorensen1987marching}. Next, we deform $\canos$ to camera view space via the deformation field $\deformfield(\canos)$. 
Then, we can check if a feature curve point $\deformfield(\curve(i))$ is occluded by the explicit mesh in view space based on z-buffer:
 \begin{equation}
     \begin{aligned}
         \label{eq:visibility}
         V_{\predc{i}} = \mathrm{zbuffer\_test}(\deformfield(\predc{i}),\deformfield(\canos)).
     \end{aligned}
 \end{equation}

However, we find that the 3D curve $\curve$ might sometimes move outside or have a scale larger than the explicit mesh $\canos$, there will be some errors if only depending on the z-buffer testing between $\predc{i}$ and $\canos$. 
We therefore make use of the SMPL surface to improve the visibility estimation, by checking 
 if the nearest point of $\predc{i}$ on the SMPL body is occluded in the camera view space in a similar way. 
Note that this is feasible as in our intersection-free curve deformation, the correspondences between $\predc{i}$ and its nearest vertice in the SMPL body are almost unchanged during optimization.
Then a curve point is considered as visible if it passes both visibility checks.

\subsection{Progressive Curve and Surface Co-evolution.}
\label{sub: surface capture}
The surface of the garment is represented by the implicit SDF.
As the feature curve visibility estimation depends on the garment surface, the curves and surface have to evolve consistently. 
To ensure the accuracy of curve visibility during the optimization process, we jointly optimize the curves and surface while imposing a regularization that the curves lie on the zero-isosurface of the SDF. 
The implicit surface is minimized by the photometric loss based on differentiable surface rendering.

\paragraph{Curve-aware Surface Initialization.} 
A good initialization for the implicit SDF $\surf{}(\msdfws{})$ can reduce the optimization difficulty and improve the performance, especially for the long skirt and dress.
Thanks to our curve-aware garment representation, we can utilize the initialized feature curve $\scaleline$ computed in ~\eref{eq:initial_curve} to enable a better shape initialization.
Specifically, we apply a handle-based deformation~\cite{sorkine2004laplacian} to deform a surface template such that its feature curves are aligned with $\scaleline$. Then, we apply IGR~\cite{gropp2020implicit} to initialize the implicit surface $\surf{}(\msdfws{})$ by fitting the deformed template.

\paragraph{Differentiable Surface Rendering.}
To reconstruct high-fidelity geometry, following the SelfRecon~\cite{jiang2022selfrecon}, we find the intersection points $\canop$ on the surface and make them differentiable~(more details can be found in supplementary).


After obtaining the intersection points $\canop$, we compute its gradient $\canon_{\canop} = \nabla f(\canop;\msdfws{})$ and transform the camera view to canonical space as $\cameraview_{\canop}$ by the Jacobian matrix of the deformed point $\deformfield(\canop)$ (more details can be found in supplementary). 
To better account for the changes in the illumination, we also take a per-frame latent code $\renderlant$ as input to the color rendering network $\rendermlp$. 
Then, the surface color $\pcolors{\canop}$ of point $\canop$ can be computed as
\begin{equation}
\begin{aligned}
    \label{eq:color_render}
    \pcolors{\canop} &= \rendermlp(\canop, \canon_{\canop}, \cameraview_{\canop}, \renderlant ,\pos{\canop};\renderws). 
\end{aligned}    
\end{equation}

\subsection{Loss Function}
\label{sub: loss}
 
The overall loss function consists of two parts, one part is for the feature curves optimization and the other is for garment surfaces optimization.

\subsubsection{Explicit Feature Curve Loss}
The optimization of feature curves relies on the 2D projection loss, a curve slope regularization loss, and an on-surface regularization loss that ensures the feature curves are on the garment surface.

\paragraph{Feature Curve Projection Loss.}  
Given SMPL pose parameter $\smplweights_{i}$ and the camera projection matrix $\Pi$, we warp predicted feature curve $\curve$ to camera view space via deformation field $\deformfield{}$, and compute project loss $\lossproj$ measured by 2D visible curves $\curvetwo$ using Chamfer Distance (CD):
\begin{equation}
    \lossproj = \projf(V_{\curve}\otimes\Pi(\deformfield(\curve)), \curvetwo)
\end{equation}
where $V_{\curve}$ is the visibility mask of curves, and symbol $\otimes$ indicates the mask selection operator. 

\paragraph{Feature Curve Slope Regularization.} To maintain the curvature of 3D curve $\curve$, we design a slope loss $\lossslop$ to regularize that the slope is consistent between adjacent points
\begin{equation}
 \begin{aligned}
    \lossslop = \sum_{i=1}^{N_p}(1-cos<\mathbf{s}_{i+1}, \mathbf{s}_{i}>)
 \end{aligned}   
\end{equation}
where $\mathbf{s}_{i} = \predc{i+1} - \predc{i}$, $N_p$ is the point number in the curve, and $cos<>$ is the cosine similarity function.

\paragraph{On-surface Regularization.} 
In addition, the feature curves are required to be on the corresponding garment surface.
Hence, we introduce an as near as possible loss $\lossanap$ as:
\begin{equation}
    \lossanap = \sum_{i=1}^{N_p}|f(\predc{i}; \msdfws{})|
\end{equation}
The overall explicit feature curve loss can be written as:
\begin{equation}
\begin{split}
    \losscurve = \lprojw \lossproj + \lslopw \lossslop 
    + \lanap \lossanap
\end{split}
\end{equation}
where $\lprojw, \lslopw$ and $\lanap$ are loss weights.


\subsubsection{Garment Surface Loss} 
For a monocular video with $N_i$ frames, the learnable parameter in implicit surface reconstruction is denoted as $\surfW$:
\begin{equation}
\begin{aligned}
    \surfW &= \{\msdfws{}, \defws, \renderws\} \cup \{\deflant_{i}, \renderlant_{i} | i=1,\dots, N_i\} \\ 
\end{aligned}
\end{equation}

\paragraph{Surface Rendering Loss.} 
For a pixel within the garment mask, we compute the ray's intersection point $\canop$ on the canonical surface $\surf{}(\msdfws{})$  and apply surface rendering network to predict the color $\pcolors{\canop}$ (see~\eref{eq:color_render}).
Then the photometric loss can be computed as
\begin{equation}
    \lossrgb = \frac{1}{|\mathcal{R}|}\sum_{\canop \in \mathcal{R}}|\pcolors{\canop}(\surfW) - I_{\canop}|,
\end{equation}
where $\mathcal{R}$ is the sample point set, $I_{\canop}$ is the corresponding ground-truth pixel color from the input images.

\paragraph{Mask-guided Implicit Consistency Loss.} 
To better optimize implicit surface, following SelfRecon~\cite{jiang2022selfrecon}, we periodically extract explicit surface meshes $\canos$ in canonical space from SDF $f$ and use a differentiable renderer~\cite{wiles2020synsin} to iteratively optimize $\canos$ by a mask loss using the surface mask. 
Then the updated explicit surface $\updatesur$ will be used to supervise the implicit SDF $f$ as
\begin{equation}
    \lossconss = \frac{1}{|\updatesur|}\sum_{\canop\in \updatesur}|f(\canop; \msdfws{})|.
\end{equation}

\paragraph{Curve-guided Implicit Consistency Loss.}
We find that the explicit mesh $\updatesur$  updated by the mask loss might contain holes or even collapse in some surface areas, which will harm the learning of implicit surface (see in Fig.~\ref{fig:curve_consistency}).
To address this issue, we design an explicit curve and surface consistency loss. 
Specifically, for a specific feature curve $\curve$ that belongs to two implicit surfaces~(\eg, waist curve belongs to both the upper-clothing and bottom-clothing), we generate its closed surface $\curvesurf$ and then sample $N_a$ points from $\curvesurf$ to constrain the implicit SDF $f$ as
\begin{equation}
    \lossconsc = \frac{1}{|\curvesurf|}\sum_{\canop \in \curvesurf}|f(\canop; \msdfws{})|.
\end{equation}


\begin{figure*}[tb] \centering
    \includegraphics[width=\textwidth]{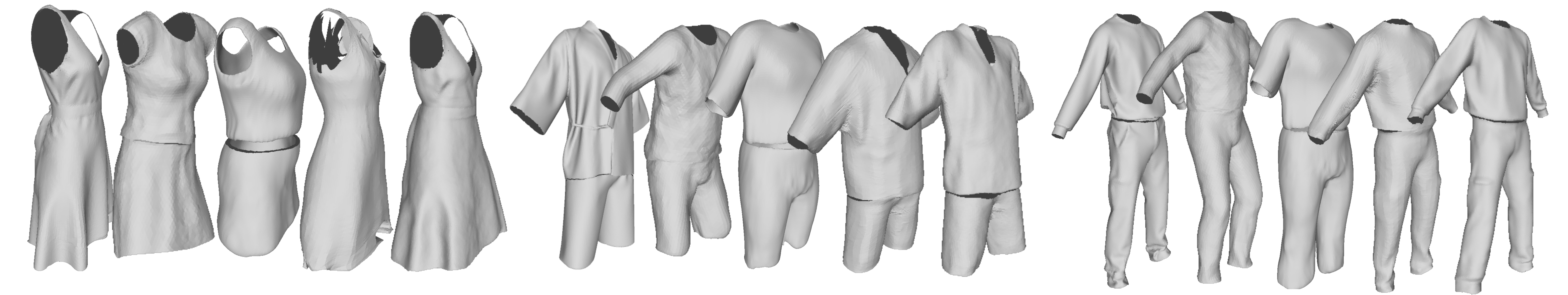}
    \caption{Qualitative comparison on the synthetic dataset. From left to right in each example: the ground-truth mesh, results of BCNet~\cite{jiang2020bcnet}, ClothedWild~\cite{moon20223d}, ReEF~\cite{zhu2022registering}, and Ours.} \label{fig:qual_synthetic}
\end{figure*}

\paragraph{Common Implicit Loss.}
Eikonal loss $\losseik$~\cite{gropp2020implicit} is included to make the implicit function the signed distance function. To avoid distortion of non-rigid transformation, a rigid loss~\cite{park2021nerfies} $\lossrigid$ is computed to constrain the non-rigid deformation.
We also compute normal loss $\lossnorm$ in canonical space to further refine the surface~\cite{jiang2022selfrecon}. 
Moreover, we compute the skeleton smoothness loss~\cite{xu2018monoperfcap} to reduce the high-frequency jitter of SMPL poses among frames~(more details can be found in supplementary).

The overall implicit surface loss can be written as:
\begin{equation}
\begin{split}
        \lossims = \lossrgb +  \lconss \lossconss + \lconsc \lossconsc\\
        \larapw \lossrigid +  \leikw \losseik  + \lnormw \lossnorm,
\end{split}
\end{equation}
where $\larapw$, $\lconss$, $\lconsc$, $\leikw$, and $\lnormw$ are the loss weights.


\begin{figure*}[tb] \centering
    \includegraphics[width=\textwidth]{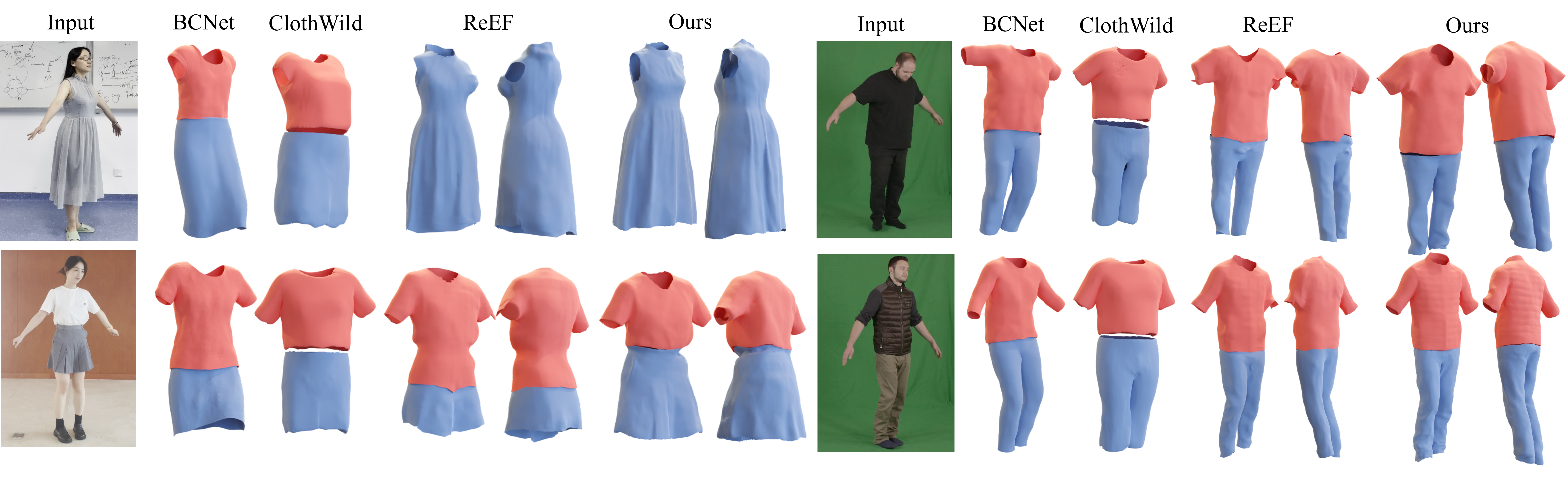}
    \vspace{-2em}
    \caption{Qualitative comparison on real datasets between BCNet~\cite{jiang2020bcnet}, ClothWild~\cite{moon20223d}, ReEF~\cite{zhu2022registering}, and our method. Upper clothes are visualized in red color, while bottom clothes and dresses are visualized in blue color. Note that BCNet and ClothWild cannot model dresses.} \label{fig:qual_real_compare}
\end{figure*}

\section{Experiments}%
\label{sec:Experiments}

Since there is no existing method for open garment meshes reconstruction from monocular videos, we compare with three state-of-the-art single-image methods, namely BCNet~\cite{jiang2020bcnet}, ClothWild~\cite{moon20223d}, and ReEF~\cite{zhu2022registering}.

\subsection{Evaluation on Synthetic Dataset}%
\label{sub:quantitative exp}
Since there is no public real dataset for evaluating dynamic garment reconstruction, we adopt four video sequences from the synthetic data generated by SelfRecon~\cite{jiang2022selfrecon} for quantitative evaluation.

We first employ Blender~\cite{blender2021} to extract the ground-truth garment mesh from the provided clothed human mesh of the first frame. 
To measure the \emph{accuracy} of the reconstructed meshes, we compute the Chamfer distance (CD) between the ground-truth and estimated meshes. 
To evaluate the \emph{temporal consistency} of the reconstructed meshes for the video sequence, we measure the consistency of corresponding vertices~(CCV), which is the root mean square error of the corresponding vertices distances in adjacent frames.

We test our method and the baseline methods on these four video sequences.
\Tref{tab:synthe results} shows that our method achieves the best results in the metrics of CD and CCV on all four videos, demonstrating the effectiveness of our method in reconstructing accurate and temporally consistent dynamic garment meshes.
From the results of high errors in the CCV, we can clearly see that single-image methods fail to maintain the consistency of the reconstruction for the video input. 
\Fref{fig:qual_synthetic} compares the visual results, in which our method produces detailed and accurate garments that are mostly close to the ground-truth surfaces.

\begin{table}[tb]\centering
    \caption{Quantitative results on four synthetic sequences.
    We compare the Chamfer distance~(CD) between the ground-truth and reconstructed surfaces~(in $cm$), as well as the consistency of corresponding vertices~(CCV) between adjacent frames.}  
    \newcommand{\chamfer}{CD}
    \newcommand{\consistency}{CCV}
    \label{tab:synthe results}
    \resizebox{0.48\textwidth}{!}{
    \begin{tabular}{*{1}{l}|*{2}{c}| *{2}{c}|*{2}{c}|*{2}{c}}
        \toprule
      & \multicolumn{2}{c|}{Female1} & \multicolumn{2}{c|}{Female3} & \multicolumn{2}{c|}{Male1} & \multicolumn{2}{c}{Male2} \\
       Method  & \chamfer & \consistency & \chamfer & \consistency & \chamfer & \consistency & \chamfer & \consistency  \\
        \midrule
        BCNet~\cite{jiang2020bcnet} & 3.184 & 7.201 & 3.447 & 6.186  
        & 2.929 & 8.604 & 5.234 & 7.128 \\
        ClothedWild~\cite{moon20223d} & 2.424 & - & 2.075 & - & 2.782 & - & 3.980 & -\\
        ReEF~\cite{zhu2022registering} & 1.810 & 3.782 & 1.924 & 4.322 & 2.005 & 6.794 &  2.865 & 3.579\\
        Ours &\Frst{1.804} & \Frst{0.597} & \Frst{1.641} & \Frst{1.064} &\Frst{1.736} & \Frst{0.484} & \Frst{1.812} & \Frst{0.433} \\
        \bottomrule
    \end{tabular}
    }

\end{table}

\begin{figure*}[tb] \centering
    \includegraphics[width=\textwidth]{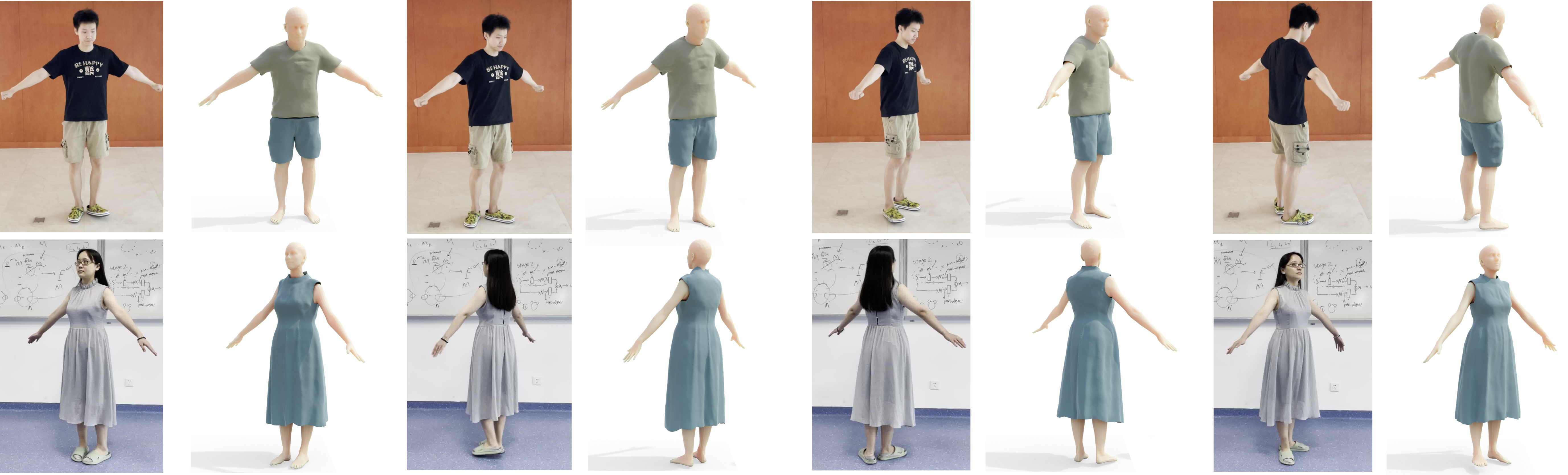}
    \caption{Dynamic garment reconstruction results of our method. Each row shows the reconstruction of four frames in a monocular video.} \label{fig:qual_real_dynamic}
\end{figure*}

\subsection{Evaluation on Real-world Videos}%
\label{sub:quantitative exp}
We then qualitatively evaluate our method on the PeopleSnapshot~\cite{alldieck2018video} and a dataset captured by ourselves. These testing videos include a diverse variety of garments categories, including upper-cloth, dress, coats, pants, and skirts.

\Fref{fig:qual_real_compare} shows the visual comparisons. The results of the baseline methods are predicted using a single image as input. 
Our method can faithfully reconstruct the layouts and surface details of the garments. In contrast, BCNet~\cite{jiang2020bcnet} and ClothWild~\cite{moon20223d} cannot accurately predict the garment layouts and produce over-smooth surfaces.

We also demonstrate our dynamic reconstruction results in
\fref{fig:qual_real_dynamic}. We can see that our method can produce space-time coherent results for different garment types (including the challenging dresses) from monocular videos, which is difficult to achieve with single-image methods.

\begin{figure}[tb] \centering
    \includegraphics[width=0.48\textwidth]{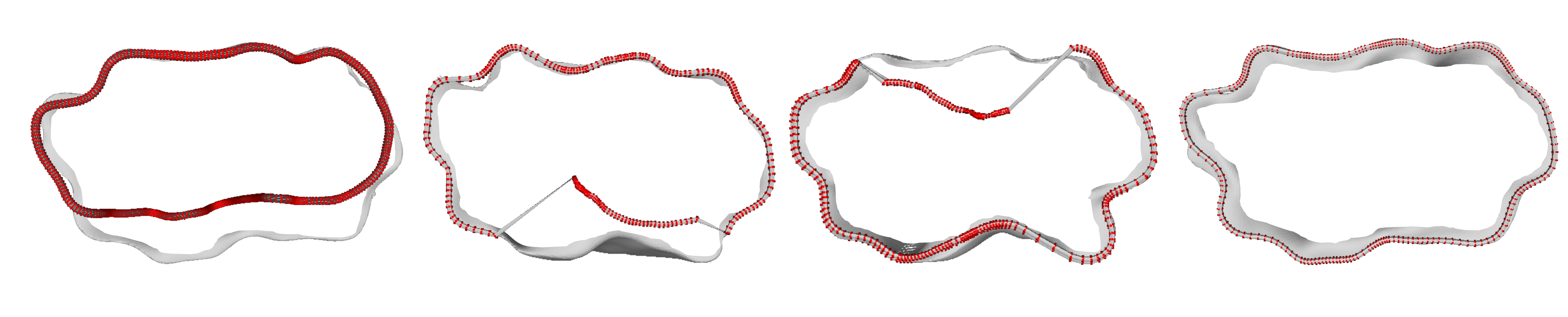}\\
    \vspace{-1em}
    \makebox[0.120\textwidth]{\scriptsize (a) Curve Normal}
    \makebox[0.120\textwidth]{\scriptsize (b) w/o Implicit Surface}
    \makebox[0.110\textwidth]{\scriptsize (c) w/o SMPL}
    \makebox[0.110\textwidth]{\scriptsize (d) Surface-aware}
    \\
    \caption{Ablation study of curve visibility estimation method.} \label{fig:abla_curve_visible}
\end{figure}

\subsection{Ablation Study}%
\label{sub:ablation_study}
We next conduct ablation study for different components of our method (more results in our supplementary material). 

\paragraph{Curve Visibility Estimation.}
As shown in \fref{fig:abla_curve_visible}, simply using normal direction for visibility estimation leads to worse results, while using both the implicit SDF and SMPL surfaces for z-buffer testing produces the best result.

\begin{figure}[tb] \centering
    \vspace{-1em}
    \includegraphics[width=0.45\textwidth]{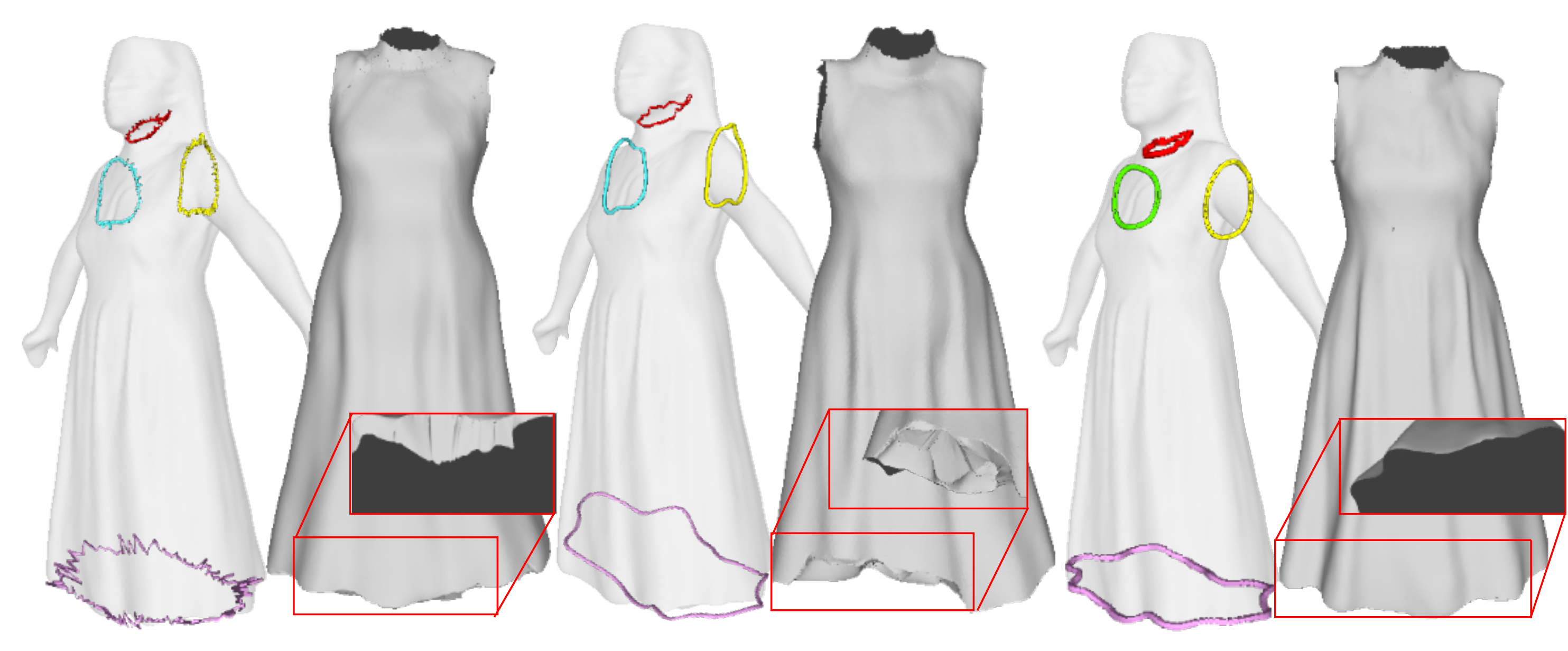}
    \\
    \vspace{-0.7em}
    \makebox[0.15\textwidth]{\footnotesize (a) w/o $\lossslop$ }
    \makebox[0.15\textwidth]{\footnotesize (b) w/o $\lossanap$ }
    \makebox[0.15\textwidth]{\footnotesize (c) Both} 
    \\
    \vspace{-0.1em}
    \caption{Ablation study of the explicit curve losses.} \label{fig:abla_curve_loss}
\vspace{0.3em}
    \includegraphics[width=0.43\textwidth]{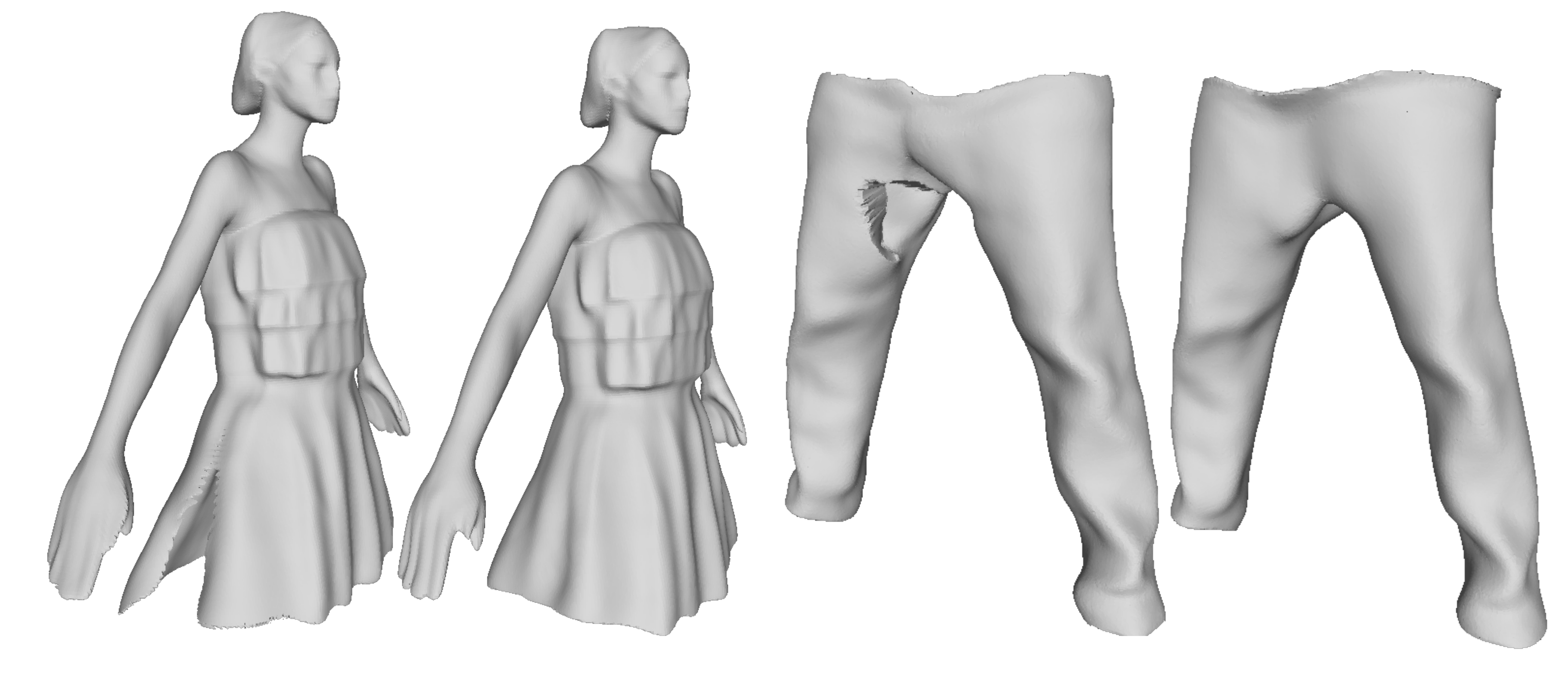} \\
    \makebox[0.10\textwidth]{\footnotesize w/o $\lossconsc$  }
    \makebox[0.10\textwidth]{\footnotesize w/ $\lossconsc$  }
    \makebox[0.10\textwidth]{\footnotesize w/o $\lossconsc$  }
    \makebox[0.10\textwidth]{\footnotesize w/ $\lossconsc$  }
    \\
    \caption{Results of w/o and w/ the curve-guided consistency loss.} \label{fig:curve_consistency}
\end{figure}

\paragraph{Explicit Curve Losses.}
\Fref{fig:abla_curve_loss}~(a) shows that without the curve slop loss $\lossslop$, the optimized curves will contain noise and artifacts.
As shown in~\fref{fig:abla_curve_loss}~(b), the proposed on-surface regularization (\ie, $\lossanap$) can well constrain the curves to be on the surface and produce much more accurate fitting results, demonstrating the implicit surface helps the optimization of curves.

\paragraph{Curve-guided Consistency Loss.}
To improve the optimization of the implicit surface, we use curves to regularize the surface. \Fref{fig:curve_consistency} shows that this regularization effectively improves the surface geometry, verifying that curves benefit the optimization of surfaces.

\section{Conclusion}%
\label{sec:Conclusion}
We have presented a new framework for dynamic garment reconstruction from monocular videos, by 
formulating this task as an optimization problem of dynamic 3D curves and surface recovery, followed by garment template registration.
To solve this problem, we introduce a novel approach, called \methodnameabbr, to jointly optimize the curves and surface from 2D supervision in a progressive co-evolution manner.
Experimental results show that our method can reconstruct high-fidelity dynamic garments meshes with open boundaries, significantly outperforming existing methods.

\paragraph{Limitations.} 
Our method can only reconstruct common garment categories whose contours can be represented by feature curves.
Additionally, our method requires the moving person to be observed from different angles.

\paragraph{Acknowledgement.} 
The work was supported in part by NSFC with Grant No. 62293482, the Basic Research Project No.HZQB-KCZYZ-2021067 of Hetao Shenzhen-HK S\&T Cooperation Zone. It was also partially supported by Shenzhen General Project with No.JCYJ20220530143604010, the National Key R\&D Program of China with grant No.2018YFB1800800, by NSFC No.~62202409, by Shenzhen Outstanding Talents Training Fund 202002, by Guangdong Research Projects No.2017ZT07X152 and No.2019CX01X104, by the Guangdong Provincial Key Laboratory of Future Networks of Intelligence (Grant No.2022B1212010001), and by Shenzhen Key Laboratory of Big Data and Artificial Intelligence (Grant No.ZDSYS201707251409055). It was also sponsored by CCF-Tencent Open Research Fund.

\clearpage
{\small
\bibliographystyle{ieee_fullname}
\bibliography{ref}
}

\end{document}